\documentclass[10pt, a4paper]{article}
\usepackage{lrec}
\usepackage{multibib}
\newcites{languageresource}{Language Resources}
\usepackage{graphicx}
\usepackage{tabularx}
\usepackage{soul}
\usepackage{amsmath}
\usepackage{amssymb}
\usepackage{subcaption}
\usepackage{xcolor}

\usepackage{epstopdf}
\usepackage[utf8]{inputenc}

\usepackage{hyperref}
\usepackage{xstring}

\usepackage{color}

\title{On the Learnability of Concepts\\ \vspace*{.5\baselineskip} \normalfont{ With Applications to Comparing Word Embedding Algorithms}}

\name{Adam Sutton, Nello Cristianini}

\address{Intelligent Systems Laboratory, University of Bristol, Bristol, BS8 1UB, UK 
\\\{adam.sutton, nello.cristianini\}@bris.ac.uk}

\abstract{
Word Embeddings are used widely in multiple Natural Language Processing (NLP) applications. They are coordinates associated with each word in a dictionary, inferred from statistical properties of these words in a large corpus. In this paper we introduce the notion of ``concept'' as a list of words that have shared semantic content. We use this notion to analyse the learnability of certain concepts, defined as the capability of a classifier to recognise unseen members of a concept after training on a random subset of it. We first use this method to measure the learnability of concepts on pretrained word embeddings. We then develop a statistical analysis of concept learnability, based on hypothesis testing and ROC curves, in order to compare the relative merits of various embedding algorithms using a fixed corpora and hyper parameters. We find that all embedding methods capture the semantic content of those word lists, but fastText performs better than the others. \\ \newline  }

\begin{document}

\maketitleabstract

\section{Introduction}
Word embedding is a technique used in Natural Language Processing (NLP) to map a word to a numeric vector, in a way that semantic similarity between two words is reflected in geometric proximity in the embedding space. This allows NLP algorithms to keep in consideration some aspects of meaning, when processing words. Typically word embeddings are inferred by algorithms from large corpora based on statistical information. These are unsupervised algorithms, in the sense no explicit information about the meaning of words is given to the algorithm. Word embeddings are used as input to multiple downstream systems such as text classifiers \cite{tang2014learning} or machine translations \cite{cho2014learning}.

An important problem in designing word embeddings is that of evaluating their quality, since a measure of quality can be used to compare the merits of different algorithms, different training sets, and different parameter settings. Importantly, it can also be used as an objective function to design new and more effective procedures to learn embeddings from data. Currently, most word embedding methods are trained based on statistical co-occurrence information and are then assessed based on criteria that are different than the training ones.

Cosine similarity and euclidean distances have shown the ability to represent semantic relationships between words such as in GloVe where the vector representations for the words man, woman, king and queen are such that \cite{pennington2014glove}:
\begin{equation}
king-queen\approx man-woman
\end{equation}

Schnabel et al. \cite{schnabel2015evaluation} identifies two families of criteria: intrinsic and extrinsic, the first family assessing properties that a good embedding should have (eg: analogy, similarity, etc), the second assessing their contribution as part of a software pipeline (eg in machine translation). 


We propose a criterion of quality for word embeddings, and then we present a statistical methodology to compare different embeddings. The criterion would fall under the intrinsic class of methods in the classification of Schnabel et. al. \cite{schnabel2015evaluation}, and has similarities with both their coherence criterion and with their categorization and relatedness criteria. However it makes use of the notion of ``concept learnability" based on statistical learning ideas. We make use of extensional definitions of concepts, as they have been defined by \cite{anthony1997computational}.  Intuitively, a concept is a subset of the universe, and it is learnable if it is possible for an algorithm to recognise further members after learning a random subset of its members. 

The key part in this study is that of a ``concept". If the set of all words in a corpus is called a vocabulary (which can be seen as the universe), we define any subset of the vocabulary as a concept. We call a concept learnable if it is possible for a learning algorithm to be trained on a random subset of its words, and then recognise the remaining words. We argue that concept learnability captures the essence of semantic structure, and if the list of words has been carefully selected, vetted and validated by rigorous studies, it can provide an objective way to measure the quality of the embedding.

In the first experiment we will measure the learnability of Linguistic Inquiry and Word Count (LIWC) lists. We compare LIWC lists to randomly generated word lists for popular pretrained word embeddings of three different algorithms (GloVe \cite{pennington2014glove}, word2vec \cite{mikolov2013distributed}, and fastText \cite{mikolov2018advances}). We show that LIWC concepts are represented in all embeddings through statistical testing.

In our second experiment we compare the learnability of different types of embedding algorithms and settings, using a linear classifier. We compare three of these embedding methods (GloVe \cite{pennington2014glove}, word2vec \cite{mikolov2013distributed}, and fastText \cite{mikolov2018advances}) to each other. We use the same method as previous, however for this experiment we train with the same hyper parameters and corpus across all three word embeddings \cite{July2019Wiki}. We show that from this experiment fastText performs the best, performing significantly better than both word2vec and GloVe.

This study is a statistical analysis of how a given word embedding affects the learnability of a set of concepts, and therefore how well it captures their meaning. We report on the statistical significance of how learnable various concepts are under different types of embedding, demonstrating a protocol for the comparison of different settings, data sets, algorithms. At the the same time this also provides a method to measure the semantic consistency of a given set of words, such as those routinely used in Social Psychology, eg. in the LIWC technique.
\section{Related Work}
Word embedding algorithms can be generated taking advantage of the statistical co-occurrence of words, assuming that words that appear together often have a semantic relationship. Three such algorithms that take advantage of this assumption are fastText \cite{mikolov2018advances}, word2vec \cite{word2vec}, and GloVe \cite{pennington2014glove}. 

There has been a lot of work focused on providing evaluation and understanding for word embeddings. Schnabel et. al. have looked at two schools of evaluation; intrinsic and extrinsic \cite{schnabel2015evaluation}. Extrinsic evaluations alone are unable to define the general quality of a word embedding. The work also shows the impact of word frequency on results, particularly with the cosine similarity measure that is commonly used. Intrinsic methods have also had criticisms, with Faruqui et. al. calling word similarity and word analogy tasks unsustainable and showing issues with the method \cite{faruqui2016problems}.

Nematzadeh et. al. showed that GloVe and word2vec have similar constraints when compared to earlier work on geometric models \cite{nematzadeh2017evaluating}. For example, a human defined triangle inequality such as ``asteroid" being similar to ``belt" and ``belt" being similar to ``buckle" are not well represented within these geometric models.

Schwarzenberg et. al. have have defined ``Neural Vector Conceptualization" as a method to interpret what samples from a word vector space belong to a certain concept \cite{schwarzenberg2019neural}. The method was able to better identify meaningful concepts related to words using non linear relations (when compared to cosine similarity). This method uses a multi class classifier with the Microsoft Concept Graph as a knowledge base providing the labels for training. 

Sommerauer and Fokkens have looked at understanding the semantic information that has been captured by word embedding vectors \cite{sommerauer2018firearms} using concepts provided by \cite{devereux2014centre} and training binary classifiers for these concepts. Their proposed method shows that using a pretrained word2vec model some properties of words are represented within the embeddings, while others are not. For example, functions of a word and how they interact are represented (e.g. having wheels and being dangerous), however appearance (e.g. size and colour) are not as well represented. 

\section{Methods and Resources}
\subsection{Embeddings}
A corpus $\mathbf{C}$ is a collection of documents from sources such as news articles, or Wikipedia. From $\mathbf{C}$ we can extract a set of words to be a vocabulary $\mathbf{V}$. Each document in $\mathbf{C}$ is a string of words (in which the ordering of words within the document is used as part of the embedding algorithm). With a vocabulary $\mathbf{V}$ and a corpus a function $\mathbf{\Phi}$ to be defined such that $\mathbf{\Phi} : \mathbf{V} \rightarrow \mathbb{R} ^{d}$, which is mapping every word in the vocabulary to a $d$ dimensional vector. Word vectors from a word embedding are commonly formalised as $w$.

Using an embedding method $\mathbf{\Phi}$, we will now define the action of going from words in a vocabulary to an embedded space as: $\mathbf{\Phi}(word_j \in V) = w_j \in \mathbb{R}^d$. A word vector for a given word will now be defined as $w$. Word vectors are generally normalised to unit length for measurement in word analogy or word similarity tasks:

\begin{equation}
\hat{w} = \frac{w}{||w||}
\end{equation}

\subsection{Concepts}
In this paper we make use of the notion of a `concept' defined as any subset of the vocabulary, that is a set of words. Sometimes we will use the expression ``list of words", for consistency with the literature in social psychology, but we will never make use of the order in that list, so that we effectively use ``list" as another expression for ``set", in this article. We define this as a set of words $\mathbf{L} \subseteq \mathbf{V}$ (or for an embedding a set of points in $\mathbb{R}^d$ such that $\Phi(\mathbf{L}) \subseteq \Phi(\mathbf{V})$). 

We use the word vectors from a word list to define this concept in an embedding. In general, a concept is defined as any subset of a set (or a ``universe"). We would normally define a concept as an unordered list of words that have been created, validated, and understood by humans that should be learnable by machines. However for the purpose of this paper a concept can be defined as any subset of words from $\mathbf{V}$. This use is consistent with the Extensional Definition of a concept used in logic, and the same definition of concept as used in the probably approximately correct model of machine learning \cite{anthony1997computational}.
\subsection{Linear Classification \label{sec:LinearClassifier}}
A classifier is a function that maps elements of an input space (a universe, in our case a vocabulary) to a classification space. A binary linear classifier is a function that classifies vectors of a vector space $R^d$ into two classes, as follows:
\begin{equation}
\label{eqn:classifierDef}
f : \mathbf{R}^d \rightarrow \{0,1\}, f(x) = \sigma(\langle x , w \rangle + b)
\end{equation}
We will learn linear classifiers from data, using the Perceptron Algorithm on a set of labeled data, which is a set of vectors labeled as belonging to class 1 or class 0. As we will learn concepts formed by words, and linear classifiers only operate on vectors, we will apply them to the vector space generated by the word embedding, as follows. 

A linear classifier is a simple supervised machine learning model used to classify membership of an input. We will use a single layer perceptron with embeddings as input to see if it is possible for a perceptron to predict half of a word list, while being trained on its other half.

Given a word list $\mathbf{L}$ such that $\Phi(\mathbf{L}) \subseteq \Phi(\mathbf{V}) \subseteq \mathbb{R}^d$ we will define the words from this list as $\mathbf{L}^c = \mathbf{V} \setminus \mathbf{L}$. We will use $\mathbf{L}$ and $\mathbf{L}^c$ to define a train set and test set for our perceptron. We will first uniformly sample half of the words of $\mathbf{L}$, we will then sample in equal amount from $\mathbf{L^c}$. We will then append these two word lists to make $\mathbf{L_{train}}$. To produce a test set $\mathbf{L_{test}}$ we will take the remaining words that haven't been sampled from $\mathbf{L}$, and sample the same number of words again from from $\mathbf{L^c}$.

A member of the training set can be defined as $l_i\in\Phi(\mathbf{L_{train}})$. We define our prediction function $\hat{y}$ as:
\begin{equation}
\label{eqn:linearClass}
\hat{y} = \sigma((\sum_i^d \theta_i l_i ) + b)
\end{equation}
where $\theta$ and $b$ are the training parameters of the classifier and $\sigma$ is the sigmoid function. We will then train the perceptron using the cross entropy loss function:
\begin{equation}
\label{eqn:lossFn}
J = -\dfrac{1}{|\mathbf{L}|}\sum_i^{|\mathbf{L}|}y_i \log \hat{y}_i + (1 - y_i) \log (1 - \hat{y}_i)
\end{equation}
where $y_i$ is the correct class of the training sample.

\subsection{Linguistic Inquiry and Word Count}
\begin{table}
\begin{center}
\caption{Sample words from the LIWC word lists used in experiments \label{tab:LiwcExamples}}
\begin{tabular}{@{} c c c @{}} 
Full Name & Sample Words & List Name \\
\hline
Positive Emotions & happy, pretty, good & posemo \\ 
Negative Emotions & hate, worthless, enemy & negemo \\  
Anger Processes & hate, kill, pissed & anger  \\
Biological Processes & eat, blood, pain & bio  \\
Relativity & area, bend, exit & relative  \\
Affective Processes & happy, ugly, bitter & affect  \\
Social Processes & talk, us, friend & social  \\
Work Concerns & work, class, boss & work  \\
Family Concerns & mom, brother, cousin & family  \\
Health Concerns & weak, heal, blind & health 
\end{tabular}
\end{center}
\end{table}

This study uses lists of words generated by the LIWC project \cite{pennebaker2001linguistic}, a long-running effort in social psychology to handcraft, vet and validate lists of words of clinical value to psychologists. They typically aim at capturing concerns, interests, emotions, topics, of psychological significance. LIWC lists are well suited to an experiment of this kind as the words within them are common and relevant to any cross-domain corpus. 

Tab.\ref{tab:LiwcExamples} shows samples of the ten word lists used in this study as well their full names, and what they will be described as when used in the context of this study. Most word lists used have hundreds of words in them. Family is the smallest word list with a total of $54$ words being used. These word samples will used to extensionally define word lists as concepts within the embedding.

\section{Measuring Performance of Linear Classifiers}
We will measure the performance of a linear classifier by using the receiver operating characteristic (ROC) curve, a quantity defined as the performance of a binary classifier as its prediction threshold is changed between the lowest probable prediction and its highest probable prediction. This curve plots the True Positive Rate (also known as the Recall) and the False Positive Rate (also known as the fall-out) at each classification threshold possible. We also show the accuracy of the classifier, and the precision. 

Our first experiment will look at the three word embedding algorithms of GloVe, word2vec, and fastText with regards to how they perform using pre-trained word embeddings readily available online. Our second experiment will compare all three algorithms performance under identical conditions, with the same training corpus and hyper parameters.

We will take the input as the embedding representations for words, and the output being a binary classification if the word belongs to that LIWC word set ($\mathbf{L}$) or not. For the training set $\mathbf{L_{train}}$, we will uniformly random sample half of the words from the list $\mathbf{L}$ we are experimenting on. We will then sample an equal number of words from $\mathbf{L^c}$. For the test set $\mathbf{L_{test}}$ we take the remaining words from $\mathbf{L}$, and again sample another equal set of negative test samples from $\mathbf{L^c}$.

We repeat this method 1,000 times, and for each iteration of this test we generate new word lists $\mathbf{L_{train}}$ and $\mathbf{L_{test}}$ each time. This method of a linear classifier has been defined in Eqn.\ref{eqn:linearClass} and Eqn.\ref{eqn:lossFn}. This experiment is performed for the 10 LIWC word lists listed in Tab.\ref{tab:LiwcExamples}. We take their average across all 1,000 iterations of the experiment we performed. 

\begin{table}
\begin{center}
\caption{Average Performance of Linear Classifiers using LIWC word lists on randomly generated word embeddings to identify members of its own set. \label{tab:ClassRandPosNeg}}
\resizebox{\columnwidth}{!}{%
\begin{tabular}{@{} c c c c c c c c c@{}} 
$\mathbf{L}$ & Size & Accuracy & Recall & FPR & Prec & AUC  \\
\hline
$\mathbf{L_{posemo}}$ & 392 & 0.500 & 0.484 & 0.488 & 0.498 & 0.495   \\ 
$\mathbf{L_{negemo}}$ & 492 & 0.502 & 0.492 & 0.486 & 0.503 & 0.505 \\  
$\mathbf{L_{anger}}$ & 184 & 0.494 & 0.487 & 0.499 & 0.494 & 0.492 \\
$\mathbf{L_{bio}}$ & 558 & 0.506 & 0.492 & 0.483 & 0.505 & 0.504 \\
$\mathbf{L_{relative}}$ & 632 & 0.500 & 0.423 & 0.423 & 0.279 & 0.503 \\
$\mathbf{L_{affect}}$ & 908 & 0.499 & 0.490 & 0.490 & 0.500 & 0.499 \\
$\mathbf{L_{social}}$ & 396 & 0.495 & 0.485 & 0.493 & 0.496 & 0.493 \\
$\mathbf{L_{work}}$ & 322 & 0.503 & 0.496 & 0.489 & 0.503 & 0.500 \\
$\mathbf{L_{family}}$ & 54 & 0.495 & 0.509 & 0.518 & 0.495 & 0.505 \\
$\mathbf{L_{health}}$ & 232 & 0.504 & 0.499 & 0.499 & 0.499 & 0.499 \\
$\mathbf{L_{random(max)}}$ & 400 & 0.57 & 0.572 & 0.4 & 0.571 & 0.566 \\
$\mathbf{L_{random(avg)}}$ & 400 & 0.496 & 0.482 & 0.490 & 0.496 & 0.493
\end{tabular}
}
\end{center}
\end{table}

\subsection{Learning Concepts from Random Embeddings \label{sec:Rand}}
In this section we will look at using concepts that are defined in Linguistic Enquiry Word Count (LIWC) \cite{pennebaker2001linguistic} word lists to see if they can be represented using randomly generated word embeddings. In this experiment we hypothesise that randomly generated word embeddings will be unable to correctly predict members of a LIWC word list that has a semantic consistency in the real world. 

The embedding algorithm in this experiment that we are using is sampling from a Gaussian distribution with a $\mu$ of $0$ and a variance of 1 ($\sim\mathcal{N}(0, 1)$). This Gaussian distribution is sampled for each dimension for each word vector within the embedding. The vocabulary $\mathbf{V}$ will be the same vocabulary as that used by GloVe's pretrained embeddings \cite{pennington2014glove}. However no corpus $\mathbf{C}$ is required for these embeddings as statistical co-occurrence from a corpus is not used.

To achieve this we will use a linear classifier training on half of a LIWC word list along with the same number of negative samples (sampled uniformly from the vocabulary $\mathbf{V}$). We will then test on the remaining words from the LIWC list, along with another equal number of samples from $\mathbf{V}$ and look at the performance of the binary classifier. This method is as described in Sec.~\ref{sec:LinearClassifier}.

As shown in Tab.~\ref{tab:ClassRandPosNeg}, randomly generated word embeddings fail to reflect word lists that have real world semantic meanings such as LIWC. All word lists perform equally as random in predicting members of the concept that it is representing. This confirms that random embeddings are unable to capture semantic information in its embedding space, confirming our hypothesis.

\subsection{Learning Concepts from GloVe, word2vec, and fastText \label{sec:LearningConcepts}}
In this section we will look at the ability of three different word embedding algorithms to capture information in word lists that reflect real world concepts.

To ensure that these metrics are statistically significant, we have created a null-hypothesis of making random concepts based on random word lists \newline ($\mathbf{L_{random}}$) and performing the same classification task on the random concept. We repeat this test 1000 times and take the best performance for each of the metrics we look at for these random lists (which will be defined as $\mathbf{L_{random(max)}}$). Of 1000 tests, we hypothesise no concept defined by a random word list outperforms any of the word lists we test on.
\begin{table}
\begin{center}
\caption{Average Performance of a Linear Classifiers using LIWC word lists on GloVe word embeddings to identify members of its own set. Random lists are also tested to obtain a p-value and compare performances. These embeddings perform better than random embeddings, after one thousand iterations and random word lists resulting in a p-value of $<0.001$ \label{tab:ClassGlovePosNeg}}
\resizebox{\columnwidth}{!}{%
\begin{tabular}{@{} c c c c c c c c c@{}} 
$\mathbf{L}$ & Size & Accuracy & Recall & FPR & Prec & AUC  \\
\hline
$\mathbf{L_{posemo}}$ & 392 & 0.915 & 0.902 & 0.079 & 0.919 & 0.964   \\ 
$\mathbf{L_{negemo}}$ & 492 & 0.913 & 0.913 & 0.085 & 0.915 & 0.965 \\  
$\mathbf{L_{anger}}$ & 184 & 0.888 & 0.880 & 0.103 & 0.896 & 0.950 \\
$\mathbf{L_{bio}}$ & 558 & 0.895 & 0.871 & 0.087 & 0.909 & 0.954 \\
$\mathbf{L_{relative}}$ & 632 & 0.937 & 0.935 & 0.059 & 0.940 & 0.979 \\
$\mathbf{L_{affect}}$ & 908 & 0.910 & 0.906 & 0.085 & 0.914 & 0.962 \\
$\mathbf{L_{social}}$ & 396 & 0.906 & 0.887 & 0.075 & 0.922 & 0.962 \\
$\mathbf{L_{work}}$ & 322 & 0.899 & 0.880 & 0.081 & 0.916 & 0.959 \\
$\mathbf{L_{family}}$ & 54 & 0.884 & 0.893 & 0.125 & 0.881 & 0.956 \\
$\mathbf{L_{health}}$ & 232 & 0.0.895 & 0.880 & 0.105 & 0.893 & 0.953 \\
$\mathbf{L_{random(max)}}$ & 400 & 0.547 & 0.32 & 0.115 & 0.617 & 0.574 \\
$\mathbf{L_{random(avg)}}$ & 400 & 0.500 & 0.198 & 0.198 & 0.502 & 0.501
\end{tabular}
}
\end{center}
\end{table}
\subsubsection{GloVe}

We will set GloVe to be our embedding algorithm ($\mathbf{\Phi}$), with the corpus $\mathbf{C}$ being a collection of Wikipedia and Gigaword 5 news articles. These embeddings are pretrained and available online on the GloVe web-page \cite{Glove6B}. These word embeddings are open for anyone to use, and can be used to repeat these experiments.

Tab.\ref{tab:ClassGlovePosNeg} shows the performance and statistics of ten different word lists from LIWC. $\mathbf{L_{random(avg)}}$ shows the average performance of concepts defined from random word lists. $\mathbf{L_{random(max)}}$ shows the best performing random word list for each test statistic.

An accuracy of approximately $0.9$ shows a high general performance. The precision and recall show that these word lists are able to accurately discern remaining members of its list and words that are not a part of the concept. After a thousand iterations of random word lists the best performing random lists (shown in $\mathbf{L_{random(max)}}$) were performing worse than each LIWC word list, giving a p-val of $<0.001$ for each word list.

\begin{table}
\begin{center}
\caption{Average Performance of Linear Classifiers using LIWC word lists on word2vec embeddings to identify members of its own set. Random lists are also tested to obtain a p-value and compare performances. word2vec is the embedding algorithm used. These embeddings perform better than random embeddings, after one thousand iterations and random word lists resulting in a p-value of $<0.001$ \label{tab:ClassW2VPosNeg}}
\resizebox{\columnwidth}{!}{%
\begin{tabular}{@{} c c c c c c c c c@{}} 
$\mathbf{L}$ & Size & Accuracy & Recall & FPR & Prec & AUC  \\
\hline
$\mathbf{L_{posemo}}$ & 392 & 0.904 & 0.914 & 0.115 & 0.888 & 0.959 \\ 
$\mathbf{L_{negemo}}$ & 492 & 0.923 & 0.920 & 0.081 & 0.919 & 0.970 \\  
$\mathbf{L_{anger}}$ & 184 & 0.890 & 0.906 & 0.126 & 0.879 & 0.953 \\
$\mathbf{L_{bio}}$ & 558 & 0.890 & 0.901 & 0.120 & 0.882 & 0.954 \\
$\mathbf{L_{relative}}$ & 632 & 0.911 & 0.952 & 0.135 & 0.876 & 0.963 \\
$\mathbf{L_{affect}}$ & 908 & 0.886 & 0.947 & 0.177 & 0.842 & 0.950 \\
$\mathbf{L_{social}}$ & 396 & 0.893 & 0.911 & 0.123 & 0.881 & 0.957 \\
$\mathbf{L_{work}}$ & 322 & 0.877 & 0.910 & 0.154 & 0.855 & 0.947 \\
$\mathbf{L_{family}}$ & 54 & 0.874 & 0.912 & 0.164 & 0.853 & 0.953 \\
$\mathbf{L_{health}}$ & 232 & 0.893 & 0.899 & 0.113 & 0.889 & 0.959 \\
$\mathbf{L_{random(max)}}$ & 400 & 0.545 & 0.27 & 0.055 & 0.68 & 0.576  \\
$\mathbf{L_{random(avg)}}$ & 400 & 0.498 & 0.128 & 0.130 & 0.494 & 0.500
\end{tabular}
}
\end{center}
\end{table}
\subsubsection{word2vec}
We will use word2vec as our embedding algorithm ($\mathbf{\Phi}$), with the corpus $\mathbf{C}$ being a dump of Wikipedia from April 2018 \cite{yamada2018wikipedia2vec} using the conventional skip-gram model. These embeddings are available online on the Wikipedia2Vec web-page \cite{yamada2018wikipedia2vec}. These word embeddings are open for anyone to use, and can be used to repeat these experiments.

Tab.~\ref{tab:ClassW2VPosNeg} shows the performance and statistics of ten different word lists from LIWC while using the word2vec embedding algorithm. $\mathbf{L_{random(avg)}}$ and \newline $\mathbf{L_{random(max)}}$ again show the average and best performances of random word lists.

An accuracy of approximately $0.9$ shows a high general performance, although it performs slightly worse than GloVe's pre-trained embeddings. This shows that the word2vec embedding algorithm $\mathbf{\Phi}$ applied to the corpus $\mathbf{C}$ yields word vectors that represent the real world meaning of words. The AUC is extracted from the scores of the sigmoid within the classifier. Overall word2vec performs slightly worse than GloVe embeddings in most metrics. However while the source corpora is very similar, GloVe has additional sources of information. The p-values for these word lists in comparison to random word lists is again $<0.001$ showing that these word lists that have a real world representation are represented accurately within the embedding. 

\begin{table}
\begin{center}
\caption{Average Performance of Linear Classifiers using LIWC word lists on fastText embeddings to identify members of its own set. Random lists are also tested to obtain a p-value and compare performances. word2vec is the embedding algorithm used. These embeddings perform better than random embeddings, after one thousand iterations and random word lists resulting in a p-value of $<0.001$ \label{tab:ClassFTPosNeg}}
\resizebox{\columnwidth}{!}{%
\begin{tabular}{@{} c c c c c c c c c@{}} 
$\mathbf{L}$ & Size & Accuracy & Recall & FPR & Prec & AUC  \\
\hline
$\mathbf{L_{posemo}}$ & 392 & 0.928 & 0.925 & 0.068 & 0.931 & 0.977 \\ 
$\mathbf{L_{negemo}}$ & 492 & 0.937 & 0.934 & 0.067 & 0.932 & 0.978 \\  
$\mathbf{L_{anger}}$ & 184 & 0.940 & 0.965 & 0.084 & 0.919 & 0.981 \\
$\mathbf{L_{bio}}$ & 558 & 0.917 & 0.933 & 0.098 & 0.905 & 0.970 \\
$\mathbf{L_{relative}}$ & 632 & 0.933 & 0.966 & 0.099 & 0.907 & 0.977 \\
$\mathbf{L_{affect}}$ & 908 & 0.886 & 0.947 & 0.177 & 0.842 & 0.950 \\
$\mathbf{L_{social}}$ & 396 & 0.927 & 0.920 & 0.074 & 0.925 & 0.973 \\
$\mathbf{L_{work}}$ & 322 & 0.918 & 0.914 & 0.077 & 0.922 & 0.970 \\
$\mathbf{L_{family}}$ & 54 & 0.966 & 0.975 & 0.041 & 0.960 & 0.995 \\
$\mathbf{L_{health}}$ & 232 & 0.931 & 0.940 & 0.078 & 0.924 & 0.980 \\
$\mathbf{L_{random(max)}}$ & 400 & 0.51 & 0.04 & 0.0 & 1.0 & 0.562  \\
$\mathbf{L_{random(avg)}}$ & 400 & 0.500 & 0.007 & 0.006 & 0.427 & 0.505 
\end{tabular}
}
\end{center}
\end{table}

\subsubsection{fastText}
The third and final word embedding algorithm ($\mathbf{\Phi}$) we will test is fastText \cite{joulin2016bag}. The corpus $\mathbf{C}$ is a collection of Wikipedia, ``UMBC WebBase corpus" and statmt.org news \cite{mikolov2018advances}. These embeddings are also pretrained word embeddings that are available from the fastText website.

Tab.~\ref{tab:ClassFTPosNeg} shows the performance statistics of the fastText word embeddings using our proposed method to evaluate word embeddings. $\mathbf{L_{random(avg)}}$ and $\mathbf{L_{random(max)}}$ show the random performance, while the other lists are LIWC word lists and their respective performances. 

A precision of 1 in the best performing random word lists are insignificant as the recall is shown to be poor, due to predicting most samples to be negative. The p-val of all of the word lists defined by LIWC is $<0.001$ as after one thousand iterations no random list outperformed any of LIWC lists. This again means that these word lists represent a real world concept, and that the embeddings are able to capture this information of this concept by using members of the set within the embedding to define it.

\subsection{Comparing Embeddings}

In this section we will be comparing the performance of the three word embedding algorithms used in the previous experiment. However, for this experiment the hyper parameters and the corpora trained will be fixed for the purpose of direct comparison. All embeddings have been generated by ourselves using the three word embedding algorithms word2vec (skip-gram), GloVe, and fastText.

The AUC metric we have previously shown can be viewed as a measure of the learnability of an embedded concept. This compares the true positive rate (also known as the recall) and the false positive rate and shows the performance at each threshold that is possible within the classifier on for a given word lists test set. 

This AUC could be seen as the performance of that binary classifier, and also as a measure of the quality of each embedding and a measure of the quality of each word list. The better the performance of an embedding, the higher perceived quality of that embedding. The better a list performs on all embeddings, the higher the quality of that list.

To accurately compare the performance of the embedding algorithms, we perform the same test as shown in Sec.\ref{sec:LearningConcepts}. However we ensure that a number of parameters are kept the same for each embedding, to maintain fairness. For this test, we will ensure that the corpus used to train will be identical between all embeddings. The corpus ($\mathbf{C}$) used for all three embedding algorithms will be a dump from the English Wikipedia taken from the first of July, 2019 \cite{July2019Wiki}. The embedding dimension $d$ will be set to 300. A word must appear a minimum of five times to be embedded, and the context window of all words is five.

In Tab.\ref{tab:AUCs} we show the AUC performance of all three embedding algorithms used in the paper. The fastText embedding algorithm is shown to have the highest performing embedding for 8 of the 10 lists that have been tested. Glove performs best on two lists, and generally performs better than word2vec overall. These performances are consistent with previous comparisons of these word embeddings \cite{pennington2014glove} \cite{mikolov2018advances}. The word list $\mathbf{L_{relative}}$ is shown to have the best overall performance across all three non-random embeddings, demonstrating the quality of that list.

\begin{table}
\begin{center}
\caption{AUC performance of word lists for each embedding algorithm used in these experiments, along with the average AUC for an embedding across all lists. Bold denotes the embedding algorithm that performs best for a given word list. Italic denotes the best performing list for each embedding algorithm.\label{tab:AUCs}}
\resizebox{\columnwidth}{!}{%
\begin{tabular}{@{} c c c c c @{}} 
$\mathbf{L}$ & GloVe & word2vec & fastText & Random \\
\hline
$\mathbf{L_{posemo}}$ & 0.961 & 0.929 & \textbf{0.965} & 0.495 \\
$\mathbf{L_{negemo}}$ & 0.965 & 0.945 & \textbf{0.973} & 0.505 \\
$\mathbf{L_{anger}}$ & 0.957 & 0.928 & \textbf{0.970} & 0.492 \\
$\mathbf{L_{bio}}$ & 0.960 & 0.935 & \textbf{0.974} & 0.504 \\
$\mathbf{L_{relative}}$ & \textbf{\textit{0.971}} & \textit{0.927} & \textit{0.961} & 0.503 \\
$\mathbf{L_{affect}}$ & \textbf{0.960} & 0.944 & 0.958 & 0.499 \\
$\mathbf{L_{social}}$ & 0.960 & 0.925 & \textbf{0.973} & 0.493 \\
$\mathbf{L_{work}}$ & 0.947 & 0.909 & \textbf{0.970} & 0.500 \\
$\mathbf{L_{family}}$ & 0.948 & 0.864 & \textbf{0.963} & 0.505 \\
$\mathbf{L_{health}}$ & 0.952 & 0.923 & \textbf{0.975} & 0.499 \\
Mean & 0.958 & 0.922 & \textbf{0.968} & 0.499  \\
Median & 0.960 & 0.927 & \textbf{0.970} & 0.499 \\
$\mathbf{L_{random(max)}}$ & 0.574 & 0.576 & 0.562 & 0.566 \\
$\mathbf{L_{random(avg)}}$ & 0.501 & 0.500 & 0.505 & 0.493
\end{tabular}
}
\end{center}
\end{table}

We tested the statistical significance of the performance differences observed between GloVe and fastText. To this purpose we performed a Wilcoxon signed-rank test, using the median of the AUCs from each embedding as the test statistic \cite{wilcoxon1992individual}. We use the Wilcoxon signed-rank test as the fastText mean AUCs shown in Tab.\ref{tab:AUCs} do not represent a normal distribution.

We propose a null hypothesis that the median difference of fastText and GloVe AUCs (as shown in Tab.\ref{tab:AUCs}) are $0$. We use a sample size of 10 as the difference of no pairs are equal to zero. We set our alpha to $0.01$ for a one sided (right) tail test, where the test statistic $W_{crit}$ is $5$. We find our resulting $W_{test}$ to be $3$, which leads us to reject the null hypothesis and show that fastText outperforming GloVe is statistically significant, for the word lists that we are testing. This gives us a p-value of $0.0088$.

\section{Conclusion}

In this paper, we have shown that word embeddings are able to capture the meaning of human defined word lists. We have shown the ability of embedding algorithms in learning concepts from word lists. In particular we have shown this quality in word2vec, GloVe and fastText. We have shown that learning embeddings from real data can represent real world concepts defined extensionally, utilising word lists provided by LIWC.

We have also shown the relative performance of GloVe, fastText, and word2vec when using LIWC word lists to form concepts using similar corpora that derive most of their corpora from Wikipedia. fastText performs better in the majority of situations for all word lists we have tested from LIWC, while GloVe outperforms word2vec generally. However as all algorithms use slightly different corpora, this result may change depending on the corpora used.

This measure of performance of word embeddings can be used in the future as a measure of ``quality" of word embeddings. While there are other methods that look at the performance of word embeddings by evaluating their performance in a specific task \cite{rajpurkar2016squad}, our method differs in that it looks at an embeddings general ability to understand human defined concepts. There has also been criticism of evaluating word embeddings using only word similarity tasks \cite{faruqui2016problems}. This method can also be used in another way as a measure of the quality of word lists and their ability to accurately describe a concept, providing an assumption or proof that an embedding is performing suitably to the users needs.

Future work with this method would involve extensive testing of the method using with varying differing hyper parameters to see the optimal performance of these embedding algorithms. An example of this is the impact of embedding dimension on performance. Another experiment could be looking at the performance of this test on deep contextualized embeddings such as ELMo \cite{peters2018deep} and BERT \cite{devlin2018bert}. These embeddings have been shown to have better performance on many tasks that employ word embeddings. While these embeddings are optimized for their specific end tasks, they train embeddings before that tuning process takes place. There is potential to compare these embeddings by testing the extracted embedding with a linear classifier, or fine tuning their full model to our task. However a key benefit for sentence embeddings is the context of words around them, which our task will not benefit from.

Further work could be focused on the performance of different word lists and concepts within word embeddings. The benefit of this could be to validate word lists that are not as carefully curated as LIWC word lists. These word lists may come from different fields, as LIWC is focused on clinical psychology other word lists may perform differently. Different source corpora may also change the performance of these word lists due to the meaning of some words changing from domain to domain. 
\section{References} 

\bibliographystyle{lrec}
\bibliography{lrec2020W-xample-kc}

\label{lr:ref}

\end{document}